\documentclass[twocolumn,compsoc,journal]{IEEEtran}
\usepackage[T1]{fontenc}
\usepackage[latin9]{inputenc}
\usepackage{float}
\usepackage{textcomp}
\usepackage{amsmath}
\usepackage{graphicx}
\usepackage[unicode=true,
 bookmarks=true,bookmarksnumbered=true,bookmarksopen=true,bookmarksopenlevel=1,
 breaklinks=false,pdfborder={0 0 0},pdfborderstyle={},backref=false,colorlinks=false]
 {hyperref}
\hypersetup{pdftitle={Your Title},
 pdfauthor={Your Name},
 pdfpagelayout=OneColumn, pdfnewwindow=true, pdfstartview=XYZ, plainpages=false}

\makeatletter

\let\SF@@footnote\footnote
\def\footnote{\ifx\protect\@typeset@protect
    \expandafter\SF@@footnote
  \else
    \expandafter\SF@gobble@opt
  \fi
}
\expandafter\def\csname SF@gobble@opt \endcsname{\@ifnextchar[
  \SF@gobble@twobracket
  \@gobble
}
\edef\SF@gobble@opt{\noexpand\protect
  \expandafter\noexpand\csname SF@gobble@opt \endcsname}
\def\SF@gobble@twobracket[#1]#2{}
\providecommand{\tabularnewline}{\\}

\usepackage[caption=false,font=footnotesize,labelfont=sf,textfont=sf]{subfig}
\usepackage[mathscr]{euscript}

\@ifundefined{showcaptionsetup}{}{%
 \PassOptionsToPackage{caption=false}{subfig}}
\usepackage{subfig}
\makeatother

\begin{document}
\title{RetinotopicNet: An Iterative Attention Mechanism Using Local Descriptors
\ \ \ \ \ \ \ \ \ \ \ \ \ \ with Global Context}
\author{Thomas~Kurbiel\ and\ Shahrzad\ Khaleghian\IEEEcompsocitemizethanks{\IEEEcompsocthanksitem Thomas~Kurbiel is with Aptiv, Wuppertal,
Germany\protect \\
e-mail: \href{http://Thomas.Kurbiel\%40aptiv.com}{Thomas.Kurbiel@aptiv.com}.

\IEEEcompsocthanksitem Shahrzad~Khaleghian is with the Department
of Marketing,\protect \\
Mercator School of Management, University Duisburg-Essen, Germany\protect 

e-mail: \href{http://Shahrzad.Kurbiel\%40uni-due.de}{Shahrzad.Kurbiel@uni-due.de}.

}}

\IEEEtitleabstractindextext{
\begin{abstract}
Convolutional Neural Networks (CNNs) were the driving force behind
many advancements in Computer Vision research in recent years. This
progress has spawned many practical applications and we see an increased
need to efficiently move CNNs to embedded systems today. However traditional
CNNs lack the property of scale and rotation invariance: two of the
most frequently encountered transformations in natural images. As
a consequence CNNs have to learn different features for same objects
at different scales. This redundancy is the main reason why CNNs need
to be very deep in order to achieve the desired accuracy.

In this paper we develop an efficient solution by reproducing how
nature has solved the problem in the human brain. To this end we let
our CNN operate on small patches extracted using the log-polar transform,
which is known to be scale and rotation equivariant. Patches extracted
in this way have the nice property of magnifying the central field
and compressing the periphery. Hence we obtain local descriptors with
global context information. However the processing of a single patch
is usually not sufficient to achieve high accuracies in e.g. classification
tasks. We therefore successively jump to several different locations,
called saccades, thus building an understanding of the whole image.
Since log-polar patches contain global context information, we can
efficiently calculate following saccades using only the small patches.
Saccades efficiently compensate for the lack of translation equivariance
of the log-polar transform.
\end{abstract}

\begin{IEEEkeywords}
convolutional neural networks, log-polar transform, spatial transformer,
scale and rotation equivariance
\end{IEEEkeywords}

}
\maketitle

\section{Introduction}

Convolutional neural networks have become ubiquitous in computer vision.
However, their complexity is quite high and usually requires expensive
GPU or FPGA implementation, which is not cost-effective for many embedded
systems. With recent industrial recognition of the benefits of artificial
neural networks for product capabilities, the demand emerged for efficient
algorithms to run in real-time on cost-effective hardware.

We identify the main reason, why traditional CNNs have to be deep
and complicated in order to achieve high accuracies, in the lacking
property of scale and rotation invariance. However scale and rotation
belong to the group of the most frequently encountered transformations
in natural images. Hence traditional CNNs have no means to generalize
features learnt on e.g. small objects to help identifying similar
objects at larger scales. Since CNNs have to relearn features for
same objects of different sizes, they require lots of training data
and lots of network parameters to do it.

In this paper we propose the use of the scale and rotation equivariant
log-polar transform, which is also utilized by nature in lower visual
areas (e.g., V1 through V5) of our brains. Furthermore humans do not
look at a scene in fixed steadiness, but instead move the eyes around,
locating informative parts crucial for understanding of the scene.
We exploit this concept of saccades to account for the missing translation
equivariance of the log-polar transform. Combining both the concept
of log-polar transform and the concept of saccades allows for a computationally
efficient implementation, whilst achieving state-of-the-art accuracies
at the same time.

\section{Related Work}

To meet the demand of running convolutional neural networks efficiently
on embedded, mobile hardware, many approaches have been persued in
the recent past. A very successful engineering oriented approach was
to directly reduce the computational cost of traditional convolutional
neural networks \cite{Freeman2018}\cite{Howard2017}.

Another approach was to directly tackle the problem of missing scale
and rotation equivariance in traditional CNNs by giving neural networks
the ability to actively spatially transform feature maps. The most
prominent representant of this family is the Spatial Transformer Network
\cite{Jaderberg2015}. Spatial Transformer helped both to achieve
higher accuracies and to reduce the computational burden.

The traditional concept of \textquotedblleft image pyramids\textquotedblright{}
allows for scale invariance, introduces however many additional parameters
and makes this approach not eligible to run in real-time on cost-effective
hardware.

The log-polar transform has in the past successfully been applied
in the classical computer vision task of image registration \cite{Wolberg2000}.
A successful coupling of log-polar sampling with state-of-the-art
deep networks has been performed in \cite{Ebel2019} to obtain richer
and more scale-invariant representations used for keypoint matching.
The authors in \cite{Esteves2018} introduce the Polar Transformer,
a special case of the Spatial Transformer that achieves rotation and
scale equivariance by using a log-polar sampling grid. In this approach
the center of the log-polar transform is determined on the whole input
image. Furthermore only a single saccade is used.

The concept of saccades has only recently been introduced for object
detection, where the authors utilize 5 keypoints as informative parts
for detection: the object center and 4 bounding box corners \cite{Lan2020}.
This approach is not operating in the log-polar space.

\section{Log-Polar Transform\label{sec:Log-Polar-Transform}}

In order to understand the log-polar transform, we briefly have to
review the concept of polar coordinates. In mathematics, the polar
coordinate system is a two-dimensional coordinate system in which
each point $(x,y)$ on a plane is represented as an angle and radius
$(\varphi,r)$, where $r$ is the distance from a reference point
$(x_{c},y_{c})$ and $\varphi$ is the angle (in degrees) from the
$x$-axis. The reference point (analogous to the origin of a Cartesian
coordinate system) is called the pole and can be chosen arbitrarily.

\begin{figure}[tbh]
\begin{centering}
\includegraphics[viewport=0bp 20bp 323bp 180bp,scale=0.5]{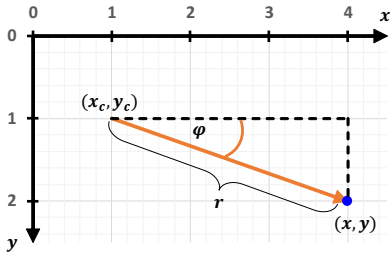}
\par\end{centering}
\caption{Point $(x,y)=(4,2)$ in Cartesian coordinates is transformed\newline$\protect\phantom{dddddd}$into
$(\varphi,r)=(18.44\text{°},3.16)$ in polar coordinates\newline$\protect\phantom{dddddd}$with
pole $(x_{c},y_{c})=(1,1)$\label{fig:polar-coordinates}}
\end{figure}

Now if take the logarithm of $r$ we arrive at the definition\textit{}\footnote{\textit{Definition in terms of Cartesian coordinates}}
of the log-polar transform:
\begin{equation}
\rho=\log r=\log\sqrt{(x-x_{c})^{2}+(y-y_{c})^{2}},
\end{equation}
\begin{equation}
\varphi=\tan^{-1}\left(\frac{y-y_{c}}{x-x_{c}}\right),
\end{equation}
The transformation from log-polar coordinates back to Cartesian coordinates
is:
\begin{equation}
x=e^{\rho}\cdot\cos\left(\varphi\right)+x_{c},\label{eq:log_polar_eqn3}
\end{equation}
\begin{equation}
y=e^{\rho}\cdot\sin\left(\varphi\right)+y_{c},\label{eq:log_polar_eqn4}
\end{equation}
In the following examples we will assume a pole coinciding with the
origin of the Cartesian coordinate system, i.e. $(x_{c},y_{c})=(0,0)$.
To gain more familiarity with the properties of the log-polar transform,
we will use an oversimplified example consisting of the following
four points:

\begin{table}[tbh]
\centering{}%
\begin{tabular}{|c|c|c|}
\cline{2-3} \cline{3-3} 
\multicolumn{1}{c|}{} & $x$ & $y$\tabularnewline
\hline 
1. & 3.0 & 3.0\tabularnewline
\hline 
2. & 4.5 & 3.0\tabularnewline
\hline 
3. & 4.5 & 4.5\tabularnewline
\hline 
4. & 3.0 & 4.5\tabularnewline
\hline 
\end{tabular}\ \ \ \ \ \ \ \ \ \ %
\begin{tabular}{|c|c|c|}
\cline{2-3} \cline{3-3} 
\multicolumn{1}{c|}{} & $\varphi$ & $\log r$\tabularnewline
\hline 
1. & 45° & 1.45\tabularnewline
\hline 
2. & 33.7° & 1.69\tabularnewline
\hline 
3. & 45° & 1.85\tabularnewline
\hline 
4. & 56.3° & 1.69\tabularnewline
\hline 
\end{tabular}
\end{table}

We will follow the convention of drawing the $\varphi$ values on
the vertical axis and the $r$ values on the horizontal axis. This
way we achieve that points lying close to the origin, will be depicted
on the left side, while points further away from the origin will be
on the right side.

\begin{figure}[tbh]
\begin{centering}
\subfloat[Cartesian coordinate system]{\includegraphics[viewport=10bp 0bp 305bp 250bp,scale=0.44]{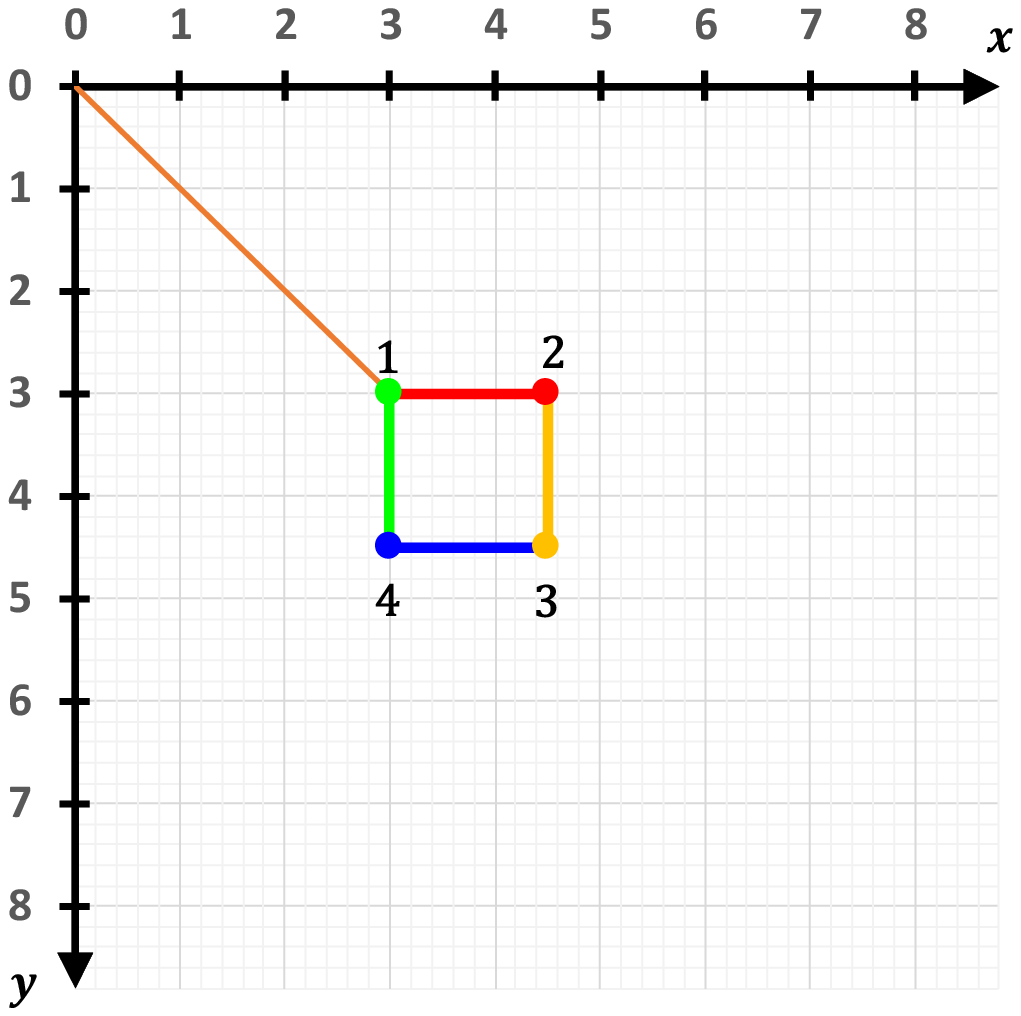}

}\subfloat[log-polar coordinate system]{\includegraphics[viewport=15bp 0bp 305bp 250bp,scale=0.44]{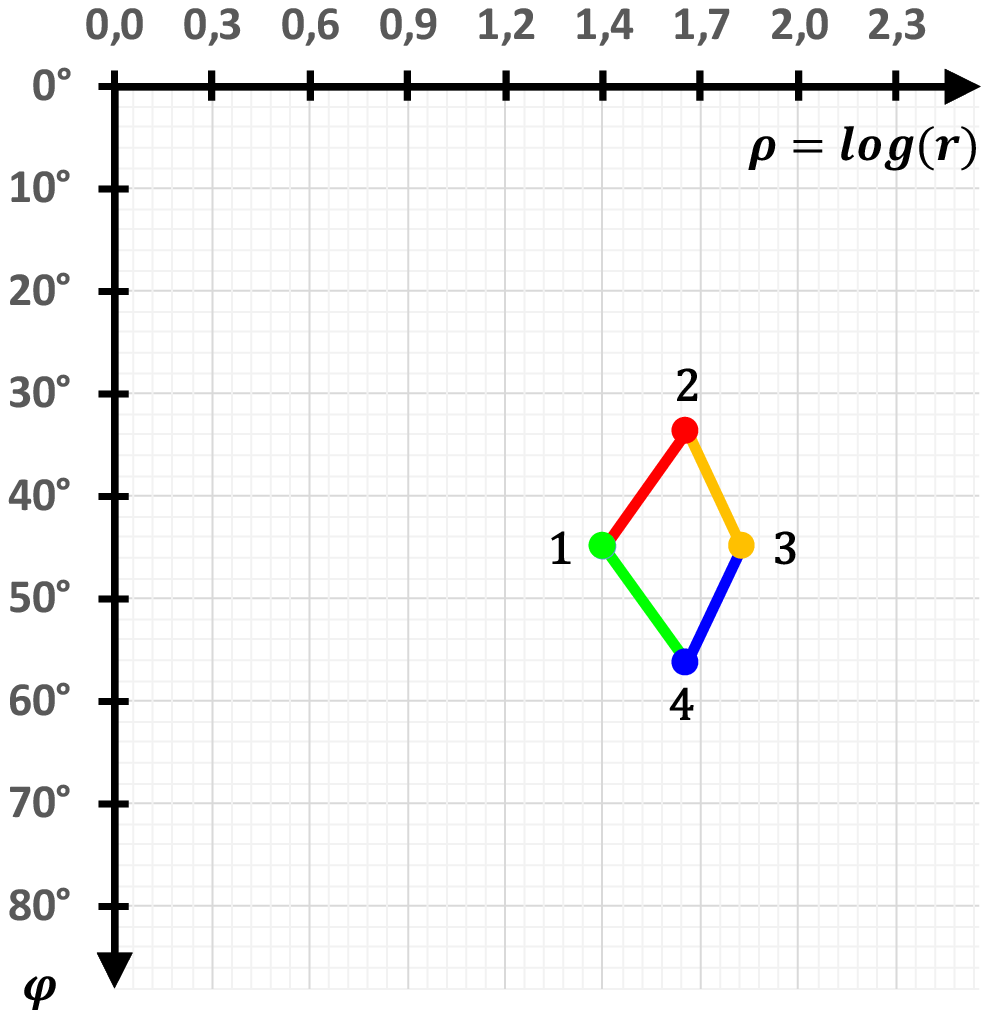}

}
\par\end{centering}
\caption{Point transformations\label{fig:points_transformations}}
\end{figure}

We see that the representation of the square in the log-polar space
is distorted. In the figure above we have connected the transformed
points by straight lines for simplicity reasons. If we actually transformed
the original lines, they would have been mapped onto curves. Since
the log-polar transform is a conformal mapping the angles between
the single curves would have been preserved \cite{Traver2010}.

What is so special about this simple transform, that nature chose
to utilize it in lower visual areas (e.g., V1 through V5) of our brains?
In these areas the neurons are organized in an orderly fashion called
topographic or retinotopic mapping, which is well approximated by
the log-polar transform \cite{Tootell1982}.

To see the properties we will first rotate the four points by 30°
clockwise in the Cartesian coordinate system. The result is depicted
below:

\begin{figure}[tbh]
\begin{centering}
\subfloat[Cartesian coordinate system]{\includegraphics[viewport=10bp 0bp 305bp 250bp,scale=0.44]{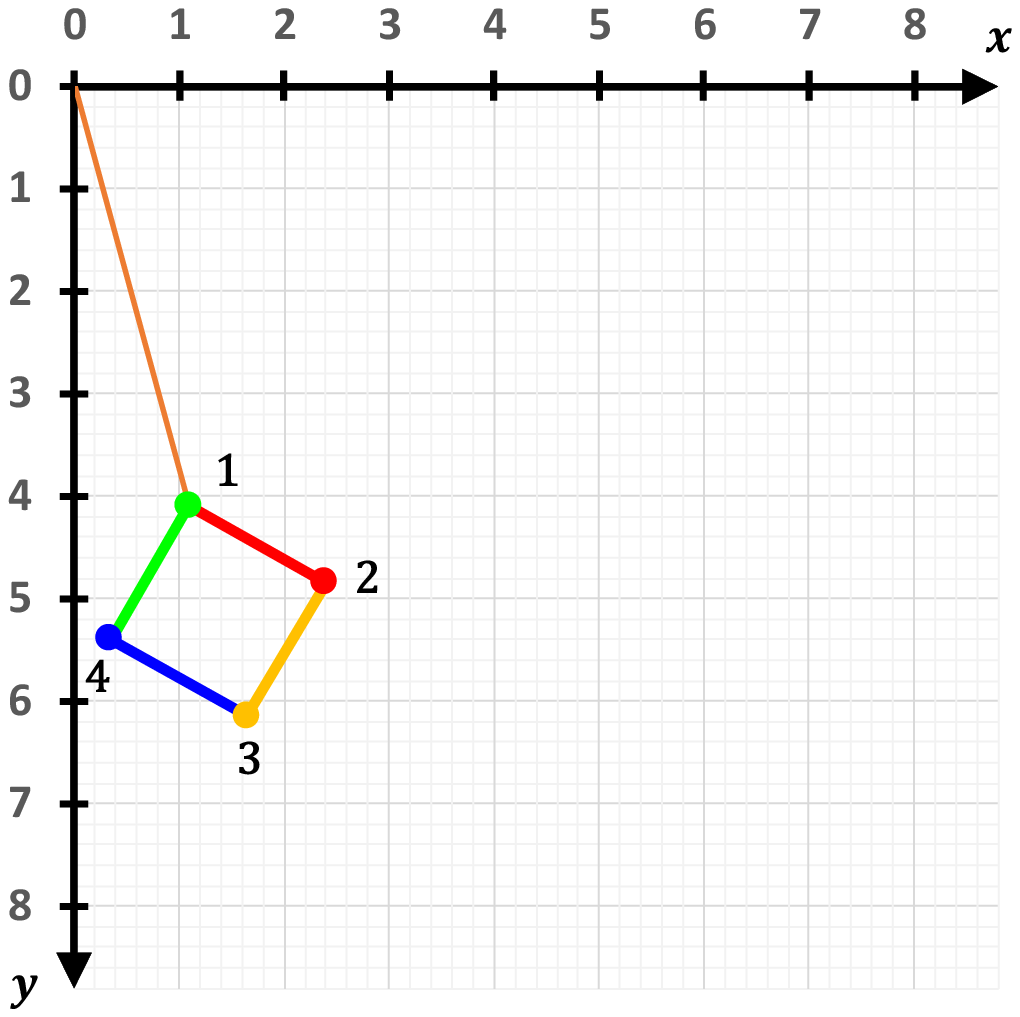}

}\subfloat[log-polar coordinate system]{\includegraphics[viewport=15bp 0bp 305bp 250bp,scale=0.44]{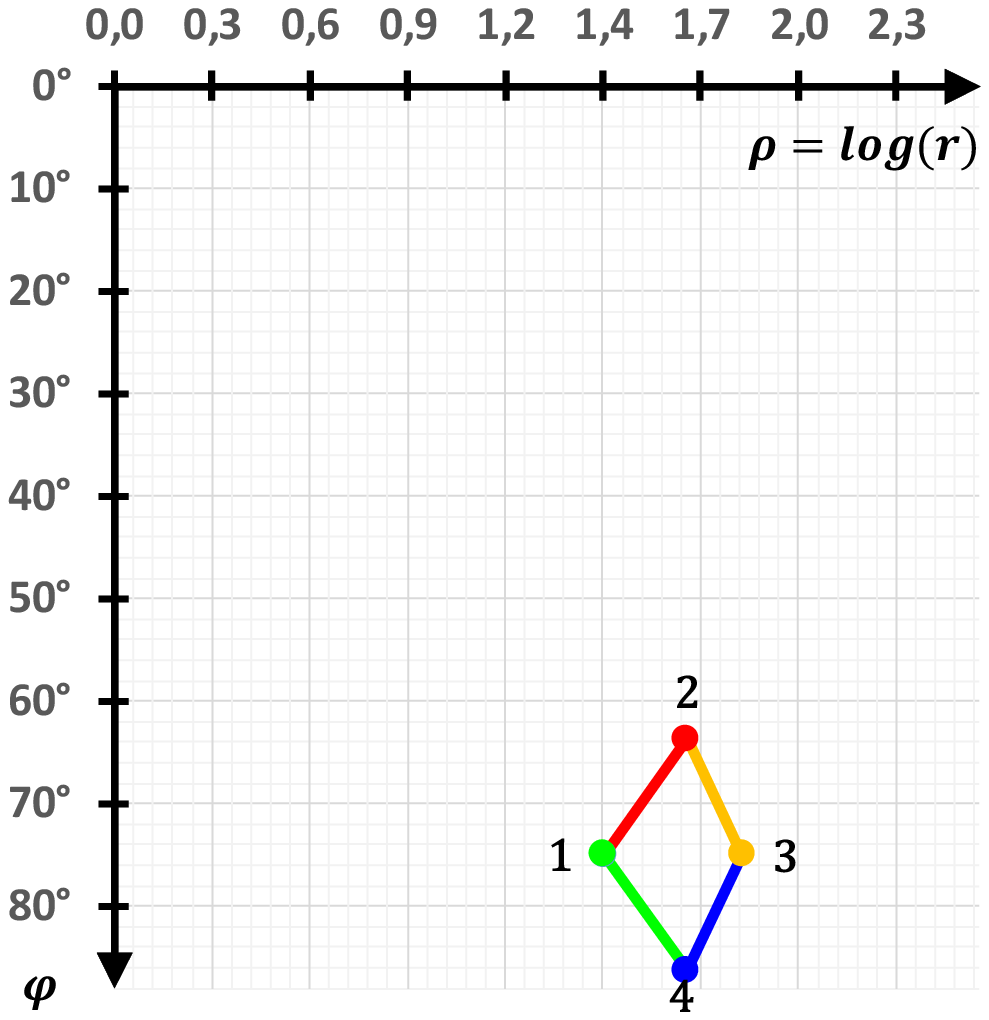}

}
\par\end{centering}
\caption{Rotation by 30° clockwise \label{fig:rotated_points_transformations}}
\end{figure}

As can be seen, the 30° rotation manifests as a circular shift along
the $\varphi$-axis in the log-polar space.

\begin{figure}[tbh]
\begin{centering}
\subfloat[Cartesian coordinate system]{\includegraphics[viewport=10bp 0bp 305bp 250bp,scale=0.44]{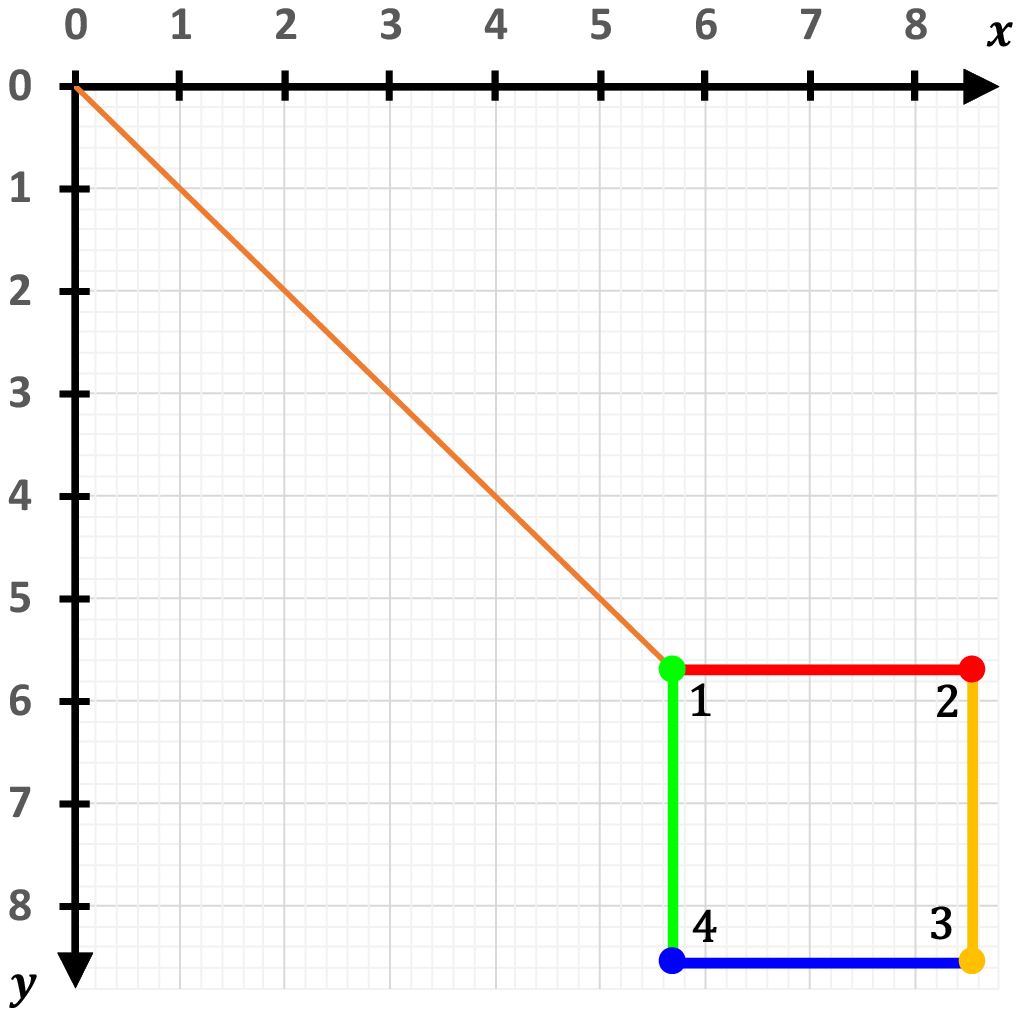}

}\subfloat[log-polar coordinate system]{\includegraphics[viewport=15bp 0bp 305bp 250bp,scale=0.44]{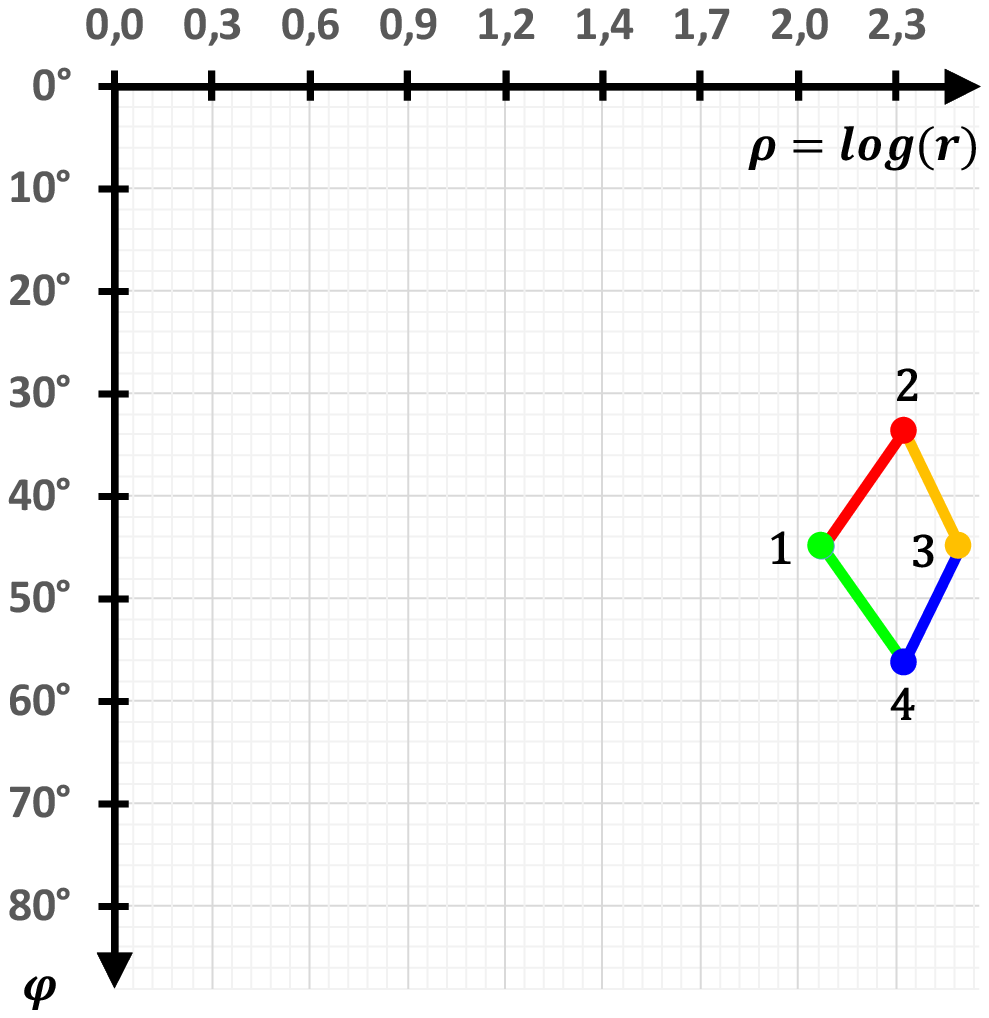}

}
\par\end{centering}
\caption{Scaling by factor of 1.9\label{fig:scaled_points_transformations}}
\end{figure}

Next we will scale all four points by the factor of 1.9. The result
is depicted in Fig.\ \ref{fig:scaled_points_transformations}. Please
note that in the Cartesian coordinate system the size of the rectangle
has almost doubled. In real world scenarios, i.e. applied on images,
that operation corresponds to a zoom of the original image by a factor
of 1.9.

\begin{figure}[tbh]
\begin{centering}
\subfloat[Input image]{\includegraphics[viewport=0bp 0bp 222bp 185bp,scale=0.44]{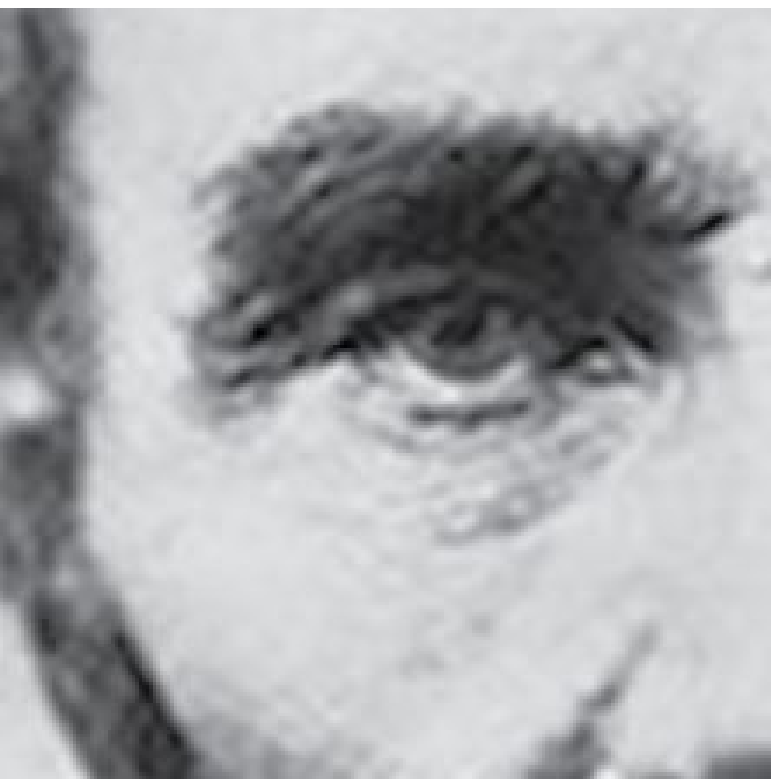}

}\subfloat[Input image\newline$\protect\phantom{ddd}$scaled 2x, rotated 45°]{\includegraphics[viewport=-20bp 0bp 222bp 185bp,scale=0.44]{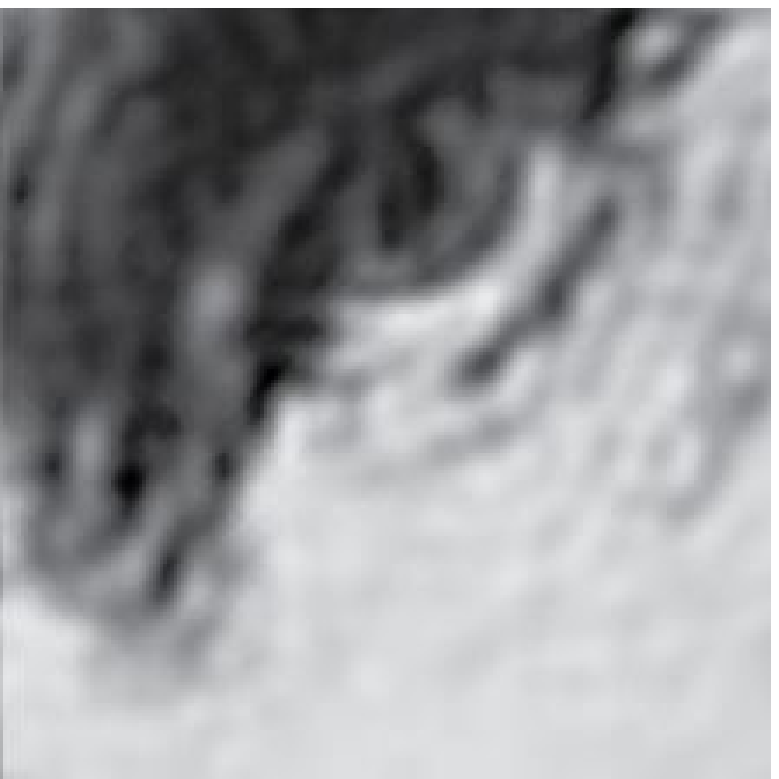}

}
\par\end{centering}
\caption{Input images in Cartesian coordinates\ \ \ \ \ \ \ \ \ \ \ \ \ \ \ \ \ \ \ (\textit{Source}:
\cite{Benton2014})\label{fig:explanation_powerful-cartesian}}
\end{figure}

As can be seen, the scaling manifests as a shift this time along the
$r$-axis in the log-polar space. This can easily also be shown mathematically:
\[
\tilde{x}=\textrm{c}\cdot x,\ \tilde{y}=\textrm{c}\cdot y
\]
\[
\tilde{\rho}=\log\sqrt{\tilde{x}^{2}+\tilde{y}^{2}}=\log\textrm{c}\cdot\sqrt{x^{2}+y^{2}}=\rho+\log\textrm{c}
\]
We have seen that the log-polar transform is equivariant to scale
and rotations, which is a very desirable property as the following
example illustrates. Imagine a traditional convolutional neural network
is fed with the two images of an eye depicted in Fig.\ \ref{fig:explanation_powerful-cartesian}.
Image (a) is used for training and image (b) is used during test time.

Even though image (b) is just a rotated and scaled version of image
(a) a convolutional neural network will have no basis of generalizing
the learnt features. Convolutional neural networks are not equivariant
to transformations such as changes in the scale or rotation of the
image.

\begin{figure}[tbh]
\begin{centering}
\subfloat[Input image]{\includegraphics[viewport=0bp 0bp 222bp 185bp,scale=0.44]{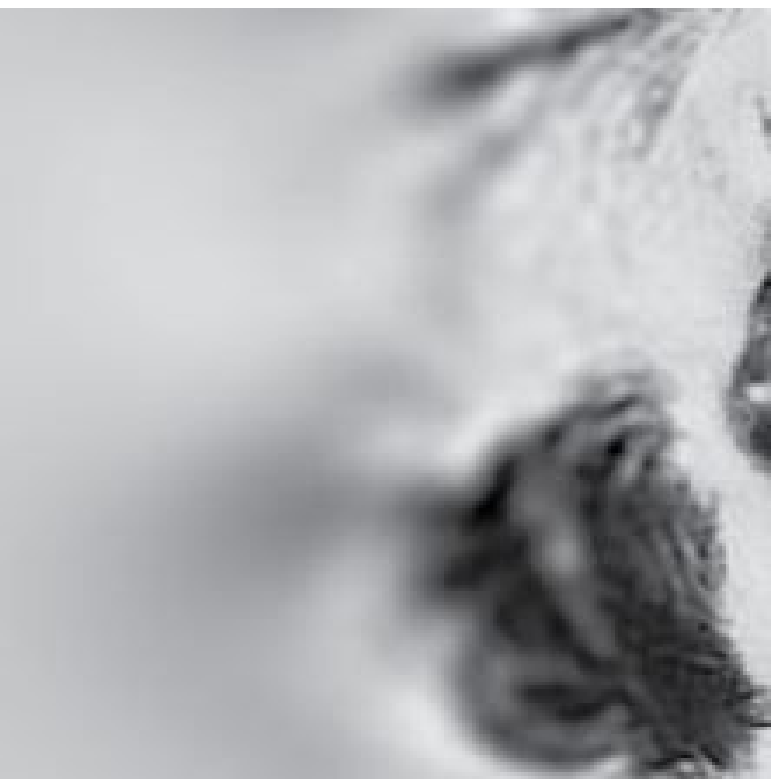}

}\subfloat[Input image\newline$\protect\phantom{ddd}$scaled 2x, rotated 45°]{\includegraphics[viewport=-20bp 0bp 222bp 185bp,scale=0.44]{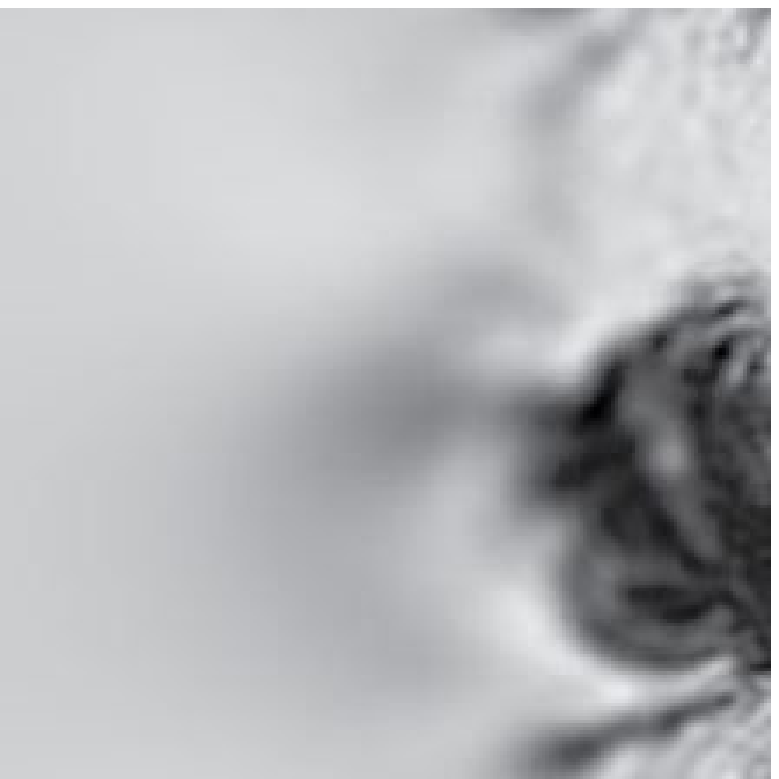}

}
\par\end{centering}
\caption{Input images in log-polar space\ \ \ \ \ \ \ \ \ \ \ \ \ \ \ \ \ \ \ \ \ \ \ \ \ \ \ (\textit{Source}:
\cite{Benton2014})\label{fig:explanation_powerful-log-polar}}
\end{figure}

If we however transform both images to the log-polar space prior to
feeding them into a convolutional neural network, than image (b) should
be a translated version of image (a), as depicted in Fig.\ \ref{fig:explanation_powerful-log-polar}.
Traditional CNNs are invariant to translations, mainly as a result
of the pooling operation, hence features learnt to classify image
(a) can now easily be reapplied for image (b). Please note that the
local context (the eyebrow) is visible in both images, while the global
context (the beard) is visible only in image (a).

\begin{figure}[tbh]
\begin{centering}
\subfloat[Cartesian coordinate system]{\includegraphics[viewport=10bp 0bp 305bp 253bp,scale=0.44]{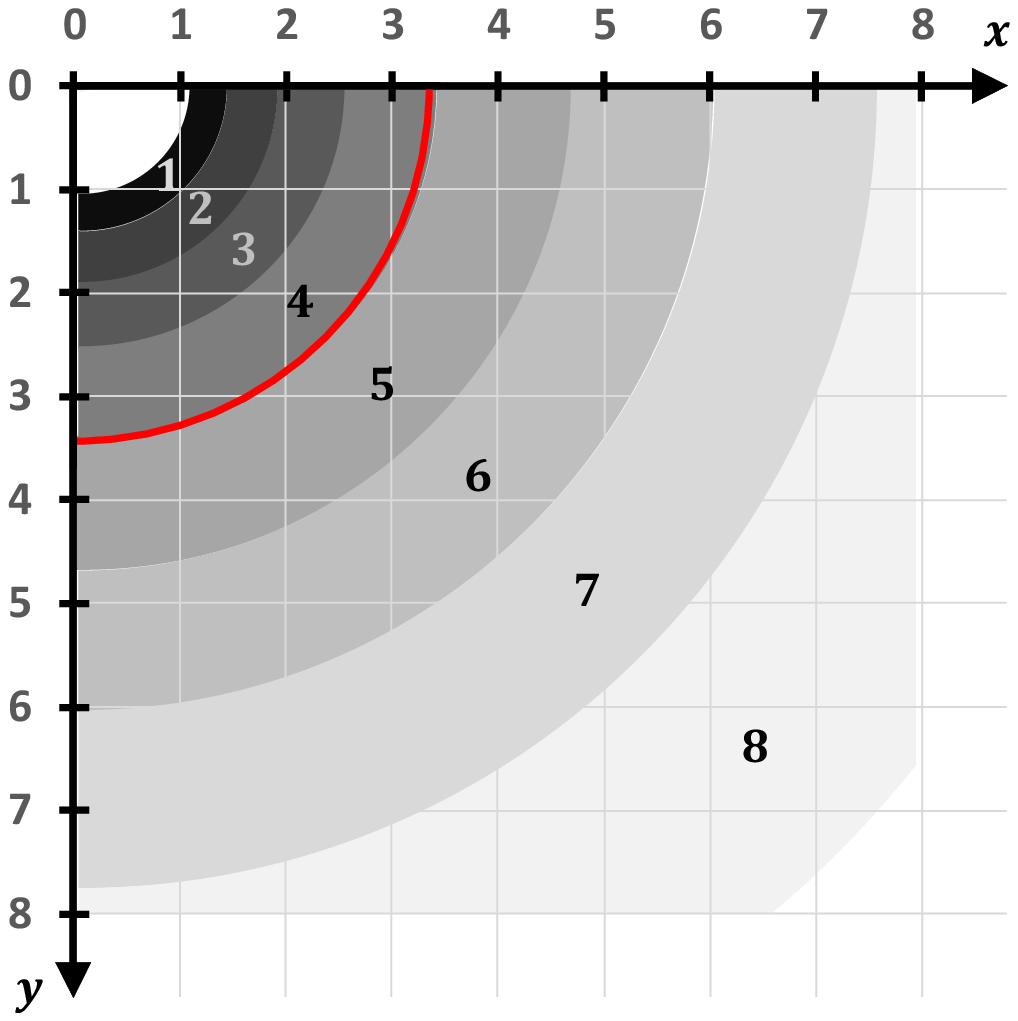}

}\subfloat[log-polar coordinate system]{\includegraphics[viewport=15bp 0bp 305bp 253bp,scale=0.44]{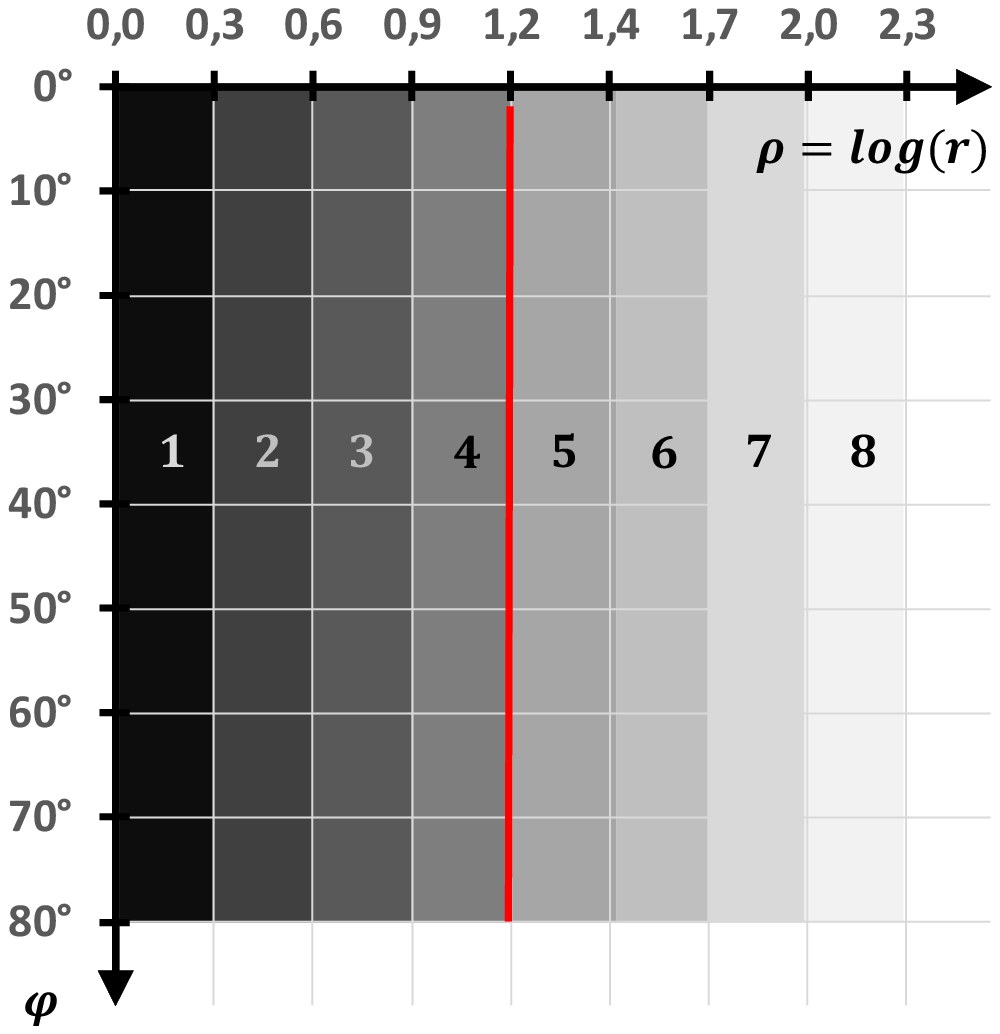}

}
\par\end{centering}
\caption{Magnification of the central field and compressesion \newline$\protect\phantom{dddddd}$of
the periphery\label{fig:compression}}
\end{figure}

Another important property of the log-polar transform is, that it
magnifies the central field and compresses the periphery, as depicted
in Fig.\ \ref{fig:compression}.

We see that the left half of the log-polar space is dedicated to a
relatively small area around the origin in the Cartesian coordinate
system. In contrast the right half of the log-polar space is dedicated
to a large area further away from the origin, called peripheral area.
We hence have high central resolution and progressively decreasing
resolution for the peripheral area. This property is the reason why
we call log-polar transformed images as local descriptors with global
context.

There is however one huge drawback of the log-polar transform, which
in the past prevented it to become more often applied: it is not translation
equivariant. Let us take the example in Fig.\ \ref{fig:points_transformations}
and shift it by 2 units to the left and by 2 units to the top.

\begin{figure}[tbh]
\begin{centering}
\subfloat[Cartesian coordinate system]{\includegraphics[viewport=10bp 0bp 305bp 253bp,scale=0.44]{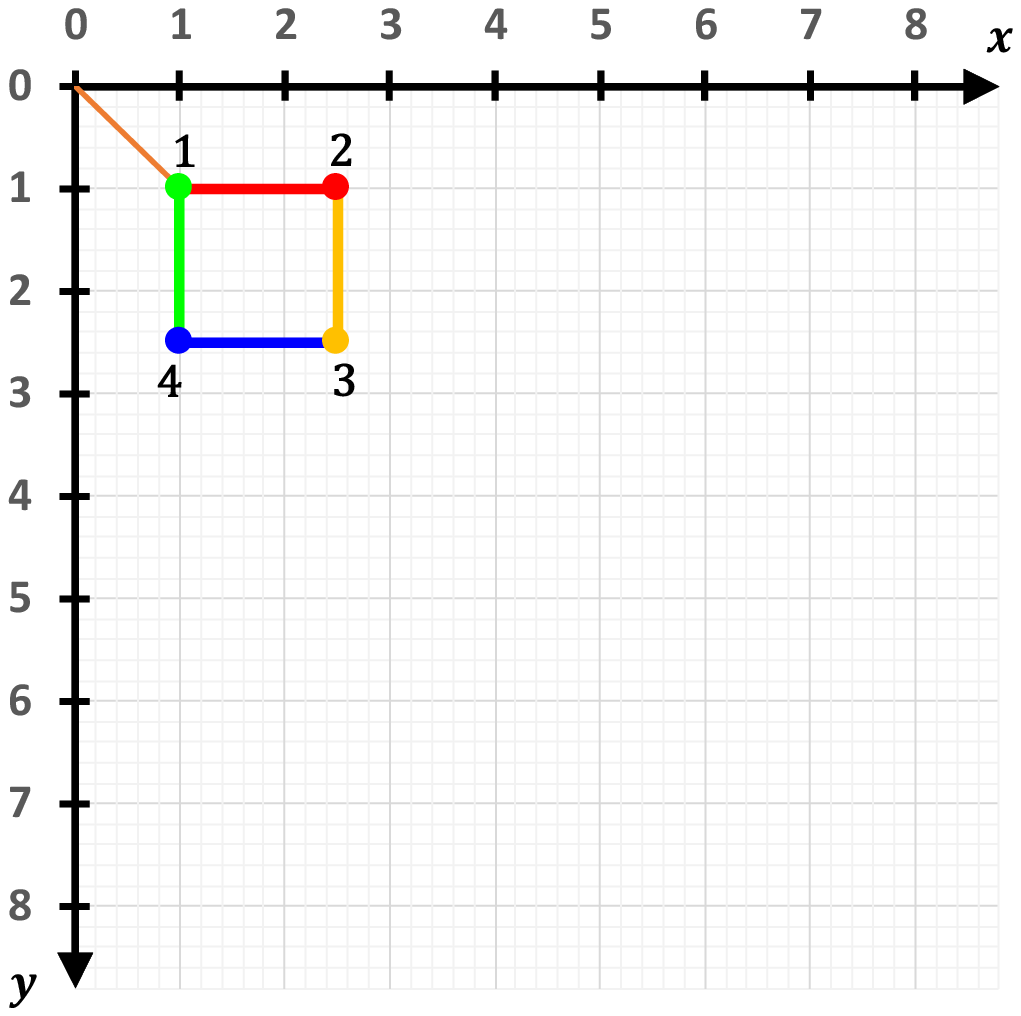}

}\subfloat[log-polar coordinate system]{\includegraphics[viewport=15bp 0bp 305bp 253bp,scale=0.44]{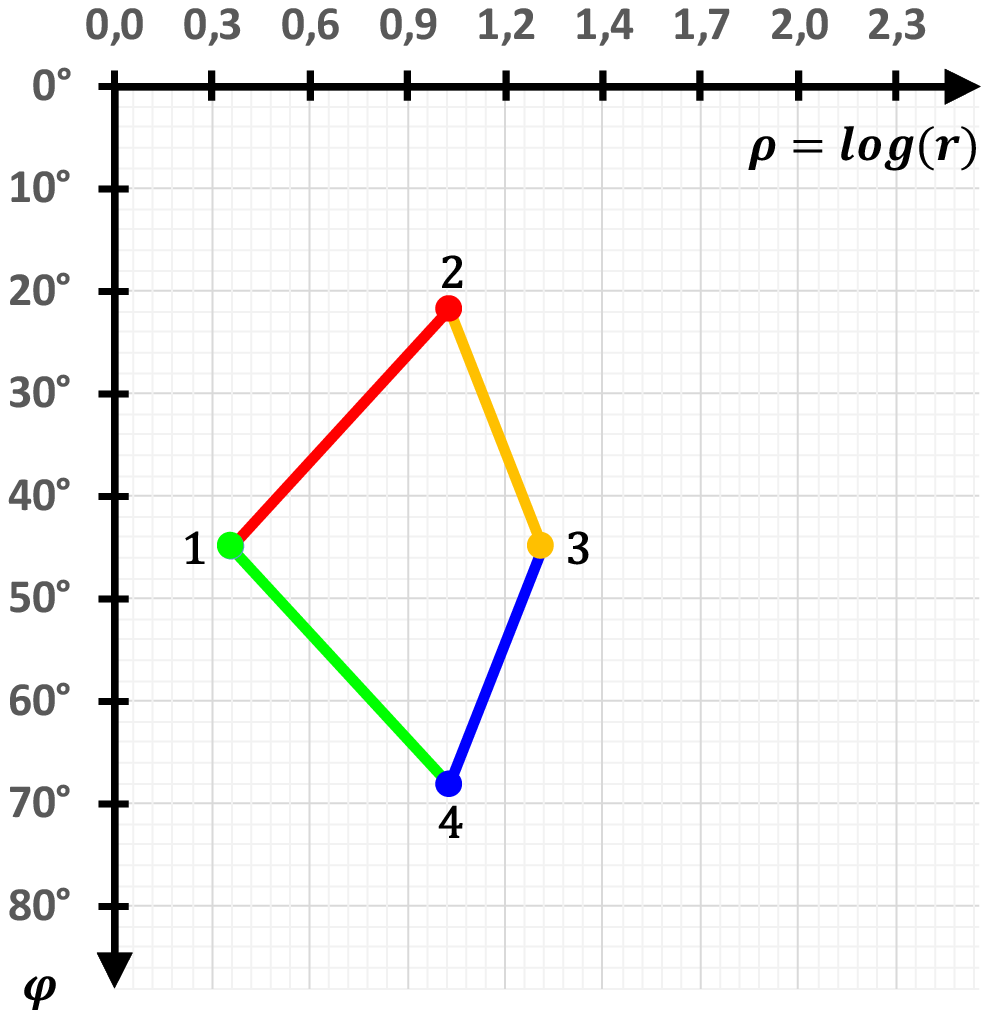}

}
\par\end{centering}
\caption{Shift to the left and to the top by 2 units\label{fig:points_transformations-1}}
\end{figure}

A rather small shift in the Cartesian coordinate system has lead to
a big distortion of the original representation of the square in the
log-polar space.

\section{Saccadic Eye Movements\label{sec:Saccadic-Eye-Movements}}

Nature has solved the shortcoming of translational equivariance by
introducing \textit{saccadic eye movement} that constantly shifts
the center of gaze from one part of the visual field to another. The
places where people fixate however are not randomly distributed. Fixations
are rather attracted by ``salient'' objects which are in the periphery
of the visual field \cite{Foulsham2008}.

\begin{figure}[tbh]
\begin{centering}
\includegraphics[viewport=0bp 18.77133bp 707bp 431.741bp,scale=0.35]{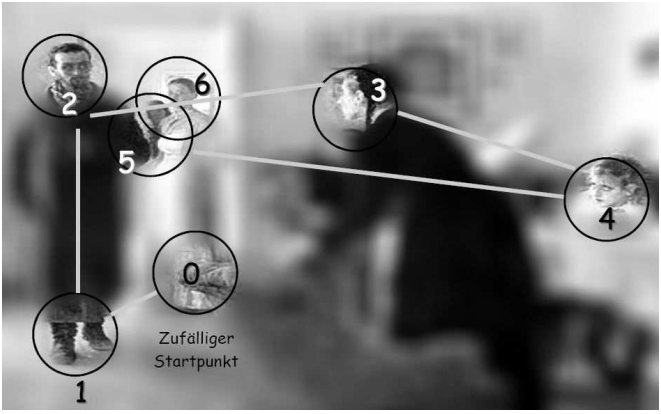}
\par\end{centering}
\caption{Saccades starting at the random initial point 0\ \ \ \ \ \ \ \ \ 
(\textit{Source:} \cite{Hunziker1967})\label{fig:saccades}}
\end{figure}

In normal viewing, several saccades are made each second and their
destinations are selected by cognitive brain process without any awareness
being involved, see Fig.\ \ref{fig:saccades}. While the eyes move
around, locating interesting parts of the scene, the brain is building
up a mental map of the scene. Hence an aggregation of previously seen
details to gain a whole understanding of the image is occurring.

In the subsequent sections, we will mimic the saccade mechanism using
tools provided by Deep Learning.

\section{Log-Polar Transform For Images}

\begin{figure}[H]
\begin{centering}
\includegraphics[viewport=0bp 30bp 397bp 160bp,scale=0.55]{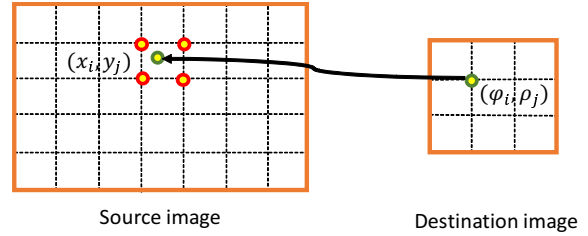}
\par\end{centering}
\caption{Image interpolation\label{fig:image_interpolation}}
\end{figure}

Transforming an image from Cartesian coordinates into the log-polar
space is performed by applying an image warping technique called reverse
mapping \cite{Gonzalez2008}. To this end the pixels in the destination
image (log-polar space) are defined to lie on a regular grid $G=\{G_{i,j}\}$
of pixels: 
\begin{gather}
G_{i,j}=(\varphi_{i},\rho_{j})\nonumber \\
=\left(i\cdot\frac{2\pi}{H'},\ \log r_{\textrm{min}}+j\cdot\frac{\log r_{\textrm{max}}-\log r_{\textrm{min}}}{W'-1}\right),
\end{gather}
with $i=0,\ldots,H'-1$ and $j=0,\ldots,W'-1$. Here $H'$ and $W'$
denote the height and width of the log-polar space, which is chosen
to be much smaller than the height $H$ and width $W$ of the source
image. Next the regular spatial grid $G$ over the destination image
is transformed back to the source image using the reverse mapping
defined in (\ref{eq:log_polar_eqn3}) and (\ref{eq:log_polar_eqn4}):
\begin{gather}
(x_{i},y_{j})=\mathcal{T}_{\theta}(G_{i,j})\nonumber \\
=\left(e^{\rho_{j}}\cdot\cos\left(\varphi_{i}\right)+x_{c},\ e^{\rho_{j}}\cdot\sin\left(\varphi_{i}\right)+y_{c}\right),\label{eq:log_polar_eqn6}
\end{gather}
where only two parameters $\theta=(x_{c},\ y_{c})$ are required,
which define the center of the patch to be transformed. Please note
that $(x_{i},y_{j})$ define the points at which the source image
is sampled. If the resulting sample points lie between pixels of the
source image, we apply bilinear interpolation using the adjacent four
pixels, depicted in Fig.\ \ref{fig:image_interpolation}.

As has been shown in \cite{Jaderberg2015} reverse mapping with bilinear
interpolation gives us a differentiable sampling mechanism, allowing
loss gradients to flow back not only to the input feature map, but
also to the sampling grid coordinates (\ref{eq:log_polar_eqn6}),
and thus back to the transformation parameters $\theta=(x_{c},\ y_{c})$.

\begin{figure}[tbh]
\begin{centering}
\includegraphics[viewport=0bp 20bp 559bp 250bp,scale=0.45]{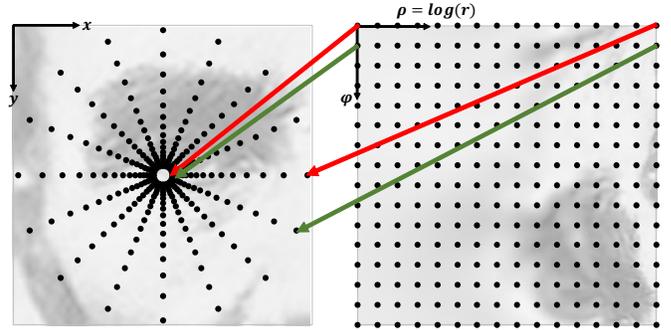}
\par\end{centering}
\caption{Sampling grid with point correspondences\label{fig:sampling_grid}}
\end{figure}

\section{Our Approach\protect\footnote{The code is available at github.com/ToKu2015/RetinotopicNet}\label{sec:Our-Approach}}

Before we dive into details of our approach, we should emphasize the
fact, that the input provided to our neural network only consists
of small patches (32x32 or 64x64 pixels) iteratively obtained by log-polar
mapping. This makes our approach computationally very efficient.

\begin{figure}[tbh]
\begin{centering}
\includegraphics[viewport=0bp 15bp 652bp 434bp,scale=0.43]{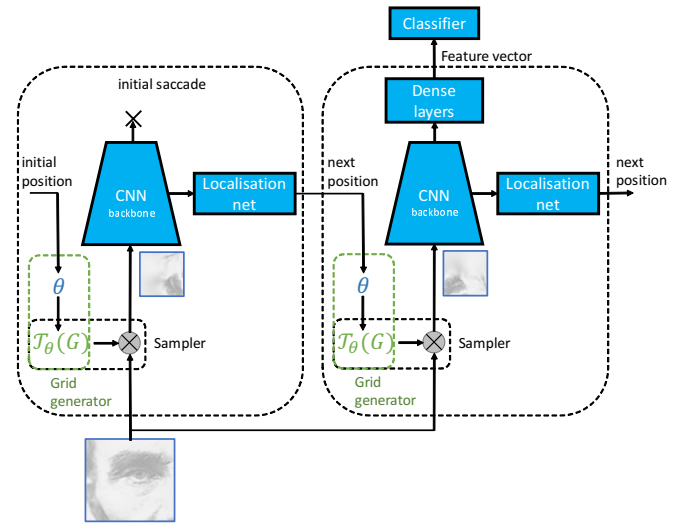}
\par\end{centering}
\caption{Basic architecture\label{fig:Basic-Architecture}}
\end{figure}

The backbone of our architecture, depicted in Fig.\ \ref{fig:Basic-Architecture},
is a small CNN network, which has the task to produce a translation
invariant feature vector representing the semantics of the log-polar
patch. The reason why the backbone is comparably small (usually consisting
of only few layers) is the following: traditional convolutional neural
networks lack the property of scale and rotation equivariance. They
have to learn different features for same objects at different scales
and rotations. This is a highly redundant task and the main reason
why for real world applications CNNs usually need to be very deep.
However when operating in log-polar space (and assuming correct positioning)
the features can be reused at different orientations and scales.,
see Sec.\ \ref{sec:Log-Polar-Transform}

In all our convolution layers we use a special kind of padding optimized
for the log-polar space. Since the log-polar space is periodic along
the $\varphi$-axis, we can use the ``wrap'' padding (cde|abcde|abc)
for this axis. For the $\rho$-axis we use the ``reflect'' padding
(cba|abcde|edc).

The second component of our architecture is a modified Spatial Transformer
Network consisting of a Grid Generator, a Sampler and a Localisation
Network \cite{Jaderberg2015}. The Grid Generator receives as only
input the reference point $(x_{c},y_{c})$ which defines the center
of the sampling grid, see Fig.\ \ref{fig:sampling_grid}. The Sampler
transforms the input image into the log-polar space. As we have discussed
previously the loss gradient can flow through the Sampler back to
the Localisation Network, hence allowing for an end-to-end learning
of the whole architecture.

\begin{figure}[tbh]
\begin{centering}
\includegraphics[viewport=0bp 30bp 738bp 380bp,scale=0.35]{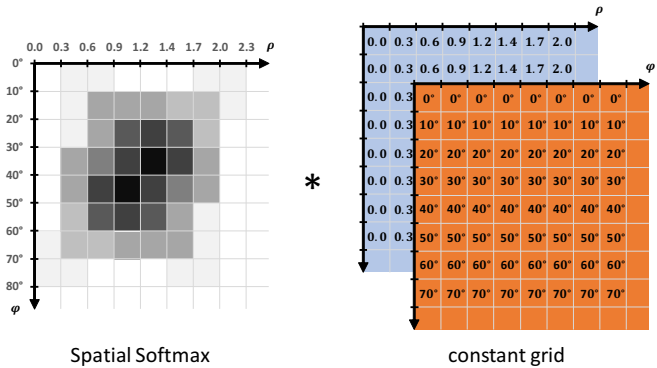}
\par\end{centering}
\caption{Last layer of the Localisation Network \label{fig:spatial_softmax}}
\end{figure}

The Localisation Network receives its input from an early layer of
the backbone CNN network, see Fig.\ \ref{fig:Basic-Architecture}.
This way a high spatial resolution is maintained, needed for identifying
interesting points for the next saccade. Besides sharing common layers
further reduces the amount of needed parameters. The Localisation
Network is designed as a fully convolutional network consisting only
of 1x1 convolution layers. The reasoning behind this design decision
is that each pixel in the feature maps of the backbone CNN network
represents a different coordinate in the log-polar space. Hence if
the network identifies a certain pixel to be interesting, it can directly
obtain its corresponding position in the log-polar space, and hence
in the Cartesian coordinate system using (\ref{eq:log_polar_eqn3})
and (\ref{eq:log_polar_eqn4}). In contrast, if we instead used a
fully connected layer this correspondence would be lost, since each
node of the fully connected layer is calculated as a weighted sum
of all spatial positions. The fully connected network would thus have
to painfully relearn an already given relation. Our numerous experiments
showed that this approach does not perform very well.

The last layer of the Localisation Network consists of a spatial softmax
with a single feature map. This feature map is multiplied (element
wise) with a constant grid representing the spatial positions of the
pixels in the log-polar space, see Fig.\ \ref{fig:spatial_softmax}.
A subsequent averaging of the product gives us the desired coordinate
of the next saccade in log-polar coordinates. Finally, to obtain the
Cartesian coordinates of the saccade, we use the reverse mapping (which
is differentiable) applying (\ref{eq:log_polar_eqn3}) and (\ref{eq:log_polar_eqn4})
and the last position (from the previous saccade).

\section{Multitask Learning}

Ideally we would like to perform only one saccade to correctly classify
an object. To achieve this goal we are training the module in Fig.\ \ref{fig:Basic-Architecture}
in a greedy manner. The whole process consists only of two iterations:
in the initial iteration we feed the network with a random initial
position (of the sampling grid) and ask the Localization Network to
determine the optimal saccade (next position). No classification is
performed yet. In the second iteration we first sample the image using
the just obtained position and then perform a classification. Needless
to say that in each iteration the same network weights are used. By
choosing a random initial position the network is forced to learn
all possible scenarios and we avoid overfitting. Using only two iterations
has furthermore the advantage of avoiding the vanishing gradient problem.
The greedy approach is our first learning task.

\begin{figure}[tbh]
\begin{centering}
\includegraphics[viewport=0bp 15bp 605bp 420bp,scale=0.43]{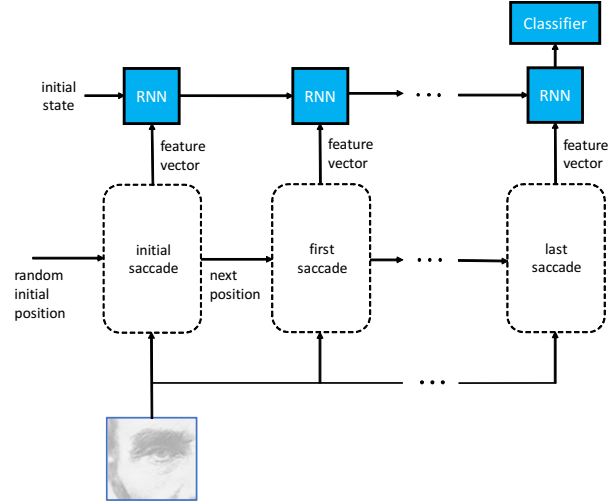}
\par\end{centering}
\caption{Aggregation of previously seen patches used to gain an \newline$\protect\phantom{ddddddd}$understanding
of the whole image\label{fig:aggregation}}
\end{figure}

However, during our experiments, we observed that often one saccade
was not sufficient to confidently classify an object. We therefore
must introduce additional (auxiliary) saccades by further unrolling
the architecture in Fig.\ \ref{fig:Basic-Architecture}. We have
mentioned in Sec.\ \ref{sec:Saccadic-Eye-Movements} that an aggregation
of previously seen patches is needed to gain an understanding of the
whole image. We achieve this by feeding the output of the last dense
layer in Fig.\ \ref{fig:Basic-Architecture} into a vanilla recurrent
neural network (RNN). The classification is performed on the output
of the RNN after the last saccade was performed. The aggregation approach
constitutes our second learning task.

\section{Evaluation}

We have performed experiments on the widely used datasets: CIFAR-10,
MNIST and Fashion-MNIST, each coming with 10 classes. In all our experiments
we have used only four saccades. The prediction made in the last saccade
is used to evaluate the performance on the test set. We have used
the following concrete implementations of the four modules depicted
in Fig.\ \ref{fig:Basic-Architecture}

\subsubsection*{CNN backbone}

The CNN backbone is a fully convolutional neural network consisting
of three convolutional layers each followed by a max-pooling layer.
We use kernel sizes of 3x3, a combination of ``wrap'' and ``reflect''
padding as described in Sec.\ \ref{sec:Our-Approach} and the ``tanh''
activation function. The sizes of feature maps are 32, 64 and 128,
respectively. The output of the last layer is flattened using average
pooling.

\subsubsection*{Classifier}

The classification net consists of three fully connected layers with
``tanh'' activation functions in the hidden layers and the ``softmax''
activation function in the last layer. The number of nodes is 128,
96, 10.

\subsubsection*{Localisation Net}

The Localisation Net get its input from the second layer of the CNN
backbone. It consists of two 1x1 convolutional layers with ``tanh''
activation function in the hidden layer and the ``spatial softmax''
activation function in the last layer. The number of nodes is 64,
64, 1. The Localisation Net is followed by the hard wired transformation
of log-polar coordinates into Cartesian coordinates, as discussed
in the last section.

\subsubsection*{RNN}

We use the vanilla version of RNN, which receives its input from the
second fully connected layer and hence has 96 nodes.

\subsection{CIFAR-10}

CIFAR-10 is a well-understood dataset and widely used for benchmarking
computer vision algorithms in the field of machine learning \cite{Krizhevsky2009}.
The dataset is comprised of 60,000 32\texttimes 32 pixel color photographs
of objects from 10 classes, such as frogs, birds, cats, ships, etc.
The dataset already has a well-defined train (50,000 examples) and
test dataset (10,000) that we have used. Top performance on the CIFAR-10
has been achieved by deep learning convolutional neural networks with
a classification accuracy above 90\% on the test dataset.

\begin{figure}[tbh]
\begin{centering}
\subfloat[Car]{\includegraphics[viewport=0bp 0bp 78bp 70.4167bp,scale=1.3]{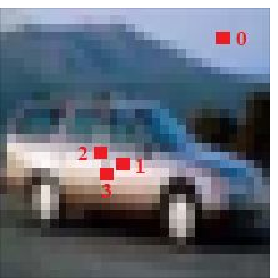}

}\subfloat[Horse]{\includegraphics[viewport=-7.22222bp 0bp 77.6389bp 70.4167bp,scale=1.3]{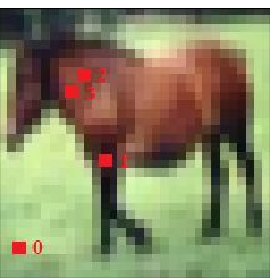}

}
\par\end{centering}
\caption{The trained network usually makes only one major saccade (a). \newline$\protect\phantom{ddddddd}$In
complicated cases several saccades are performed (b).\label{fig:saccade_cifar_10}}
\end{figure}

During training we used the following four color augmentations: hue,
saturation, brightness and contrast. In addition we use flipping and
zooming, which is a powerful augmentation that can make a network
robust to (small) changes in object size. We achieve an accuracy of
above 80\% on the test set using the small network described above.

\subsection{MNIST}

MNIST is a well-known benchmark of handwritten digit images---consisting
of a training set of 60,000 examples and a test set of 10,000 examples.
Each example is a 28x28 gray scale image, associated with a label
from 10 classes. We achieve an accuracy of above 99\% on the test
set using the small network described above.

\subsection{Fashion-MNIST}

Fashion-MNIST is a dataset of Zalando's article images. It is intended
to serve as a direct drop-in replacement for the original MNIST dataset
for benchmarking machine learning algorithms. It shares the same image
size and structure of training and testing splits. We achieve an accuracy
of above 90\% on the test set using the small network described above.

\section{Conclusion and Future Work}

In this paper we have introduced an object recognition mechanism inspired
by the human visual system. Objects are easily recognized by the human
visual system despite variation in the size of the object, its position
in the environment, or even its rotation (as in television viewing
while lying on the couch). This is mainly achieved by the use of the
scale and rotation equivariant log-polar transform. To compensate
for the lack of translation equivariance of the log-polar transform
we have introduced the concept of saccades, which allows for a guided
scanning of the whole image. The huge advantage of our approach is
that all calculations (both the classification and the calculation
of the next position to be visited) are performed on the rather small
log-polar patches. Furthermore by achieving true scale and rotation
invariance, the network can reuse features learnt for e.g. small size
objects to classify large objects. Another huge advantage is that
networks trained on small images like CIFAR-10 can still be applied
to perform classification on large images by just adapting the size
of the sampling grid.

In our future work we will concentrate on the following two topics:
we have successfully used the proposed approach for object detection.
We performed experiments on the COCO dataset using only 64x64 pixel
sized input patches. We will present our results in an upcoming paper.

Furthermore we have successfully applied the same principle of log-polar
based attention on tasks coming from the domains of speech recognition
and neural machine translation similar to the approach presented in
\cite{Luong2015}.

\bibliographystyle{plain}
\bibliography{RetinotopicNet}

\end{document}